\documentclass[a4paper,twoside]{article}

\usepackage{epsfig}
\usepackage{subcaption}
\usepackage{calc}
\usepackage{amssymb}
\usepackage{amstext}
\usepackage{amsmath}
\usepackage{amsthm}
\usepackage{multicol}
\usepackage{pslatex}
\usepackage{apalike}
\usepackage[bottom]{footmisc}
\usepackage{SCITEPRESS}     

\usepackage{cite}
\usepackage{amsmath,amssymb,amsfonts}
\usepackage{graphicx}
\usepackage{textcomp}
\usepackage{xcolor}
\usepackage{url}

\usepackage[ruled, lined, linesnumbered, commentsnumbered, longend]{algorithm2e}
\DontPrintSemicolon
\usepackage{algpseudocode}

\SetCommentSty{normalfont}
\SetKwInput{KwRequire}{Require}
\SetNlSty{}{}{:}

\usepackage{balance}
\usepackage{subcaption}
\usepackage{caption}
\usepackage{multirow}
\usepackage{booktabs}
\usepackage{array}
\usepackage{arydshln}
\usepackage{dashbox}

\usepackage{enumitem}
\usepackage{balance}

\def\BibTeX{{\rm B\kern-.05em{\sc i\kern-.025em b}\kern-.08em
    T\kern-.1667em\lower.7ex\hbox{E}\kern-.125emX}}

\usepackage{amsmath,amssymb} 
\usepackage{hyperref}
\hypersetup{colorlinks,allcolors=black}

\begin{document}

\title{Multimodal Crowd Counting with Pix2Pix GANs}

\author{\authorname{Muhammad Asif Khan\sup{1}\orcidAuthor{0000-0003-2925-8841}, Hamid Menouar\sup{1}\orcidAuthor{0000-0002-4854-909X} and Ridha Hamila\sup{2}\orcidAuthor{0000-0002-6920-7371}}
\affiliation{\sup{1}Qatar Mobility Innovations Center, Qatar University, Doha, Qatar}
\affiliation{\sup{2}Department of Electrical Engineering, Qatar University, Doha, Qatar}
\email{mkhan@qu.edu.qa, hamidm@qmic.com, hamila@qu.edu.qa}
}

\keywords{Crowd counting, CNN, density estimation, multimodal, RGB, Thermal.}

\abstract{Most state-of-the-art crowd counting methods use color (RGB) images to learn the density map of the crowd. However, these methods often struggle to achieve higher accuracy in densely crowded scenes with poor illumination. Recently, some studies have reported improvement in the accuracy of crowd counting models using a combination of RGB and thermal images. Although multimodal data can lead to better predictions, multimodal data might not be always available beforehand.
In this paper, we propose the use of generative adversarial networks (GANs) to automatically generate thermal infrared (TIR) images from color (RGB) images and use both to train crowd counting models to achieve higher accuracy. We use a Pix2Pix GAN network first to translate RGB images to TIR images. Our experiments on several state-of-the-art crowd counting models and benchmark crowd datasets report significant improvement in accuracy.}

\onecolumn \maketitle \normalsize \setcounter{footnote}{0} \vfill

\section{Introduction}

Crowd counting has gained significant attention in recent years due to its diverse applications across various domains, including crowd management, urban planning, security surveillance, event management, and public safety. The ability to accurately estimate crowd density and count individuals in congested areas holds immense practical significance, enabling better resource allocation, improved crowd control strategies, and enhanced decision-making processes. The advent of deep learning (DL) has led to a paradigm shift in crowd counting techniques, achieving higher accuracy and scalability in diverse real-world applications.

Most state-of-the-art works on crowd counting use the density estimation approach \cite{Khan2022RevisitingCC, khan2023aimag}. In density estimation, a deep learning model (typically a convolution neural network (CNN) or a vision transformer (ViT)) is trained using annotated crowd images to learn the crowd density. The annotations come in the form of sparse localization maps that indicate the head positions of individuals present in each scene or region of interest. The localization maps are converted into continuous density maps for each image by applying a Gaussian kernel centered at each head location. The intensity of the map thus represents the crowd density at different spatial locations. The density maps serve as the ground truth for the crowd images to train the crowd counting model.

A number of different models have been proposed in the last few years using the density estimation approach. These models apply various model architectural innovations and novel learning functions to improve the accuracy performance. However, most of these works use optical color images mainly with reasonable lighting conditions. However, in many surveillance scenarios images captured using optical cameras will have poor lighting conditions resulting in poor performance of the counting models. To improve the accuracy, thermal infrared (TIR) cameras are used along with the optical camera to capture both color RGB images and thermal images. Then a crowd counting model may use one (monomodal) or both (multimodal) RGB and TIR images to learn the crowd density in low-light conditions.

\begin{figure}[tp]
    \centering
    \includegraphics[width=0.4\columnwidth]{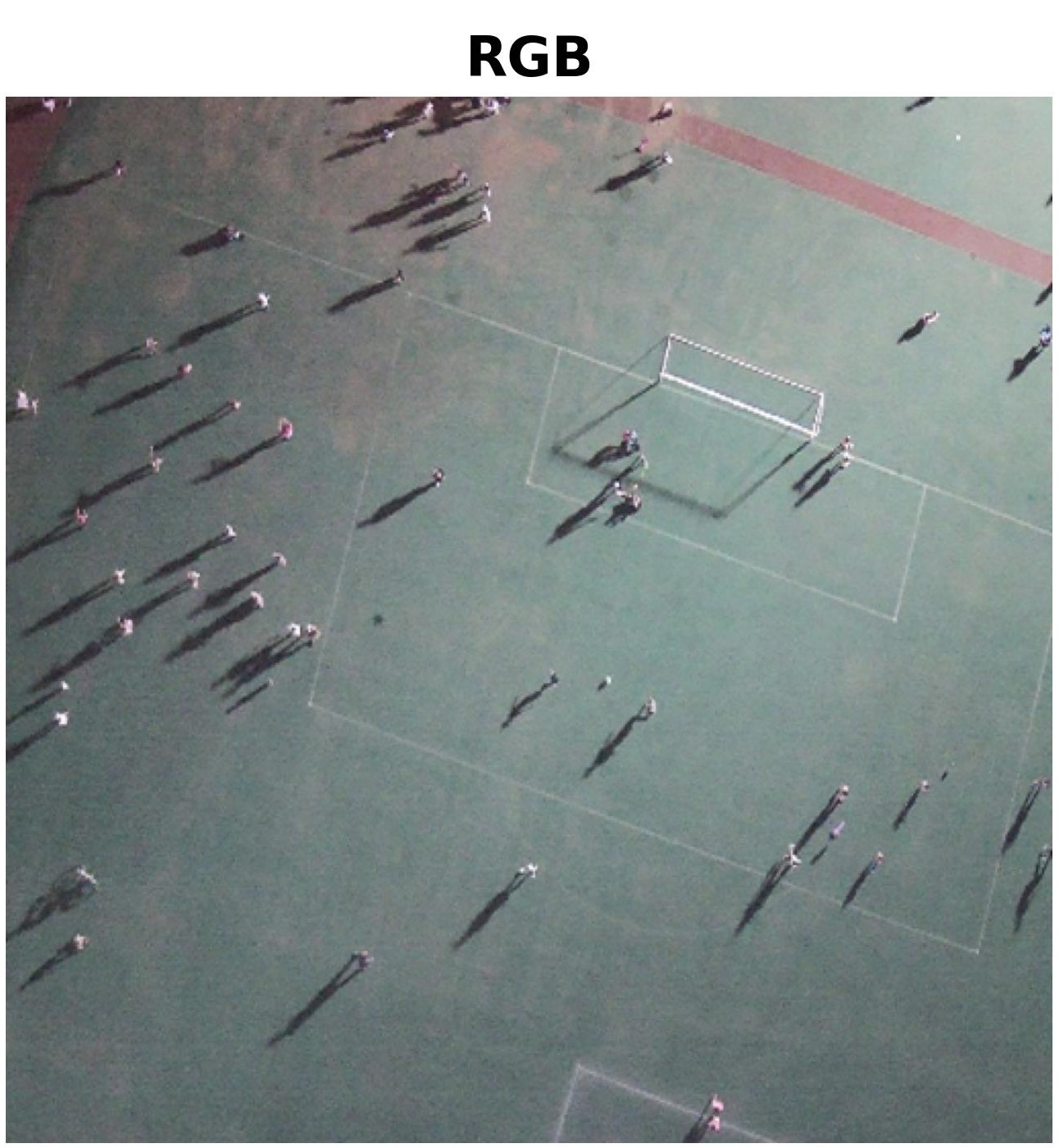} \hspace{0.1em}
    \includegraphics[width=0.4\columnwidth]{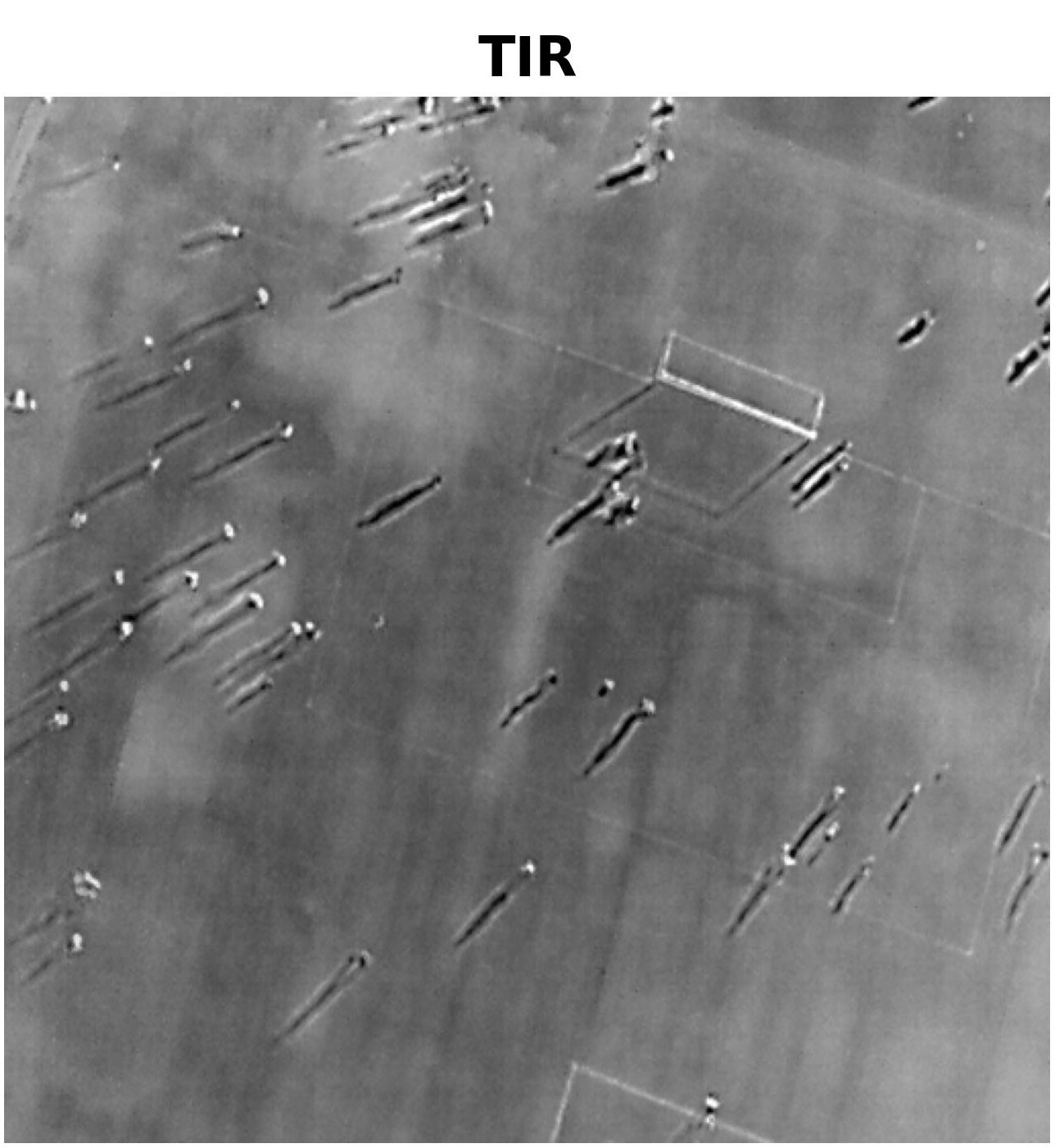} \\[0.1em]
    \includegraphics[width=0.4\columnwidth]{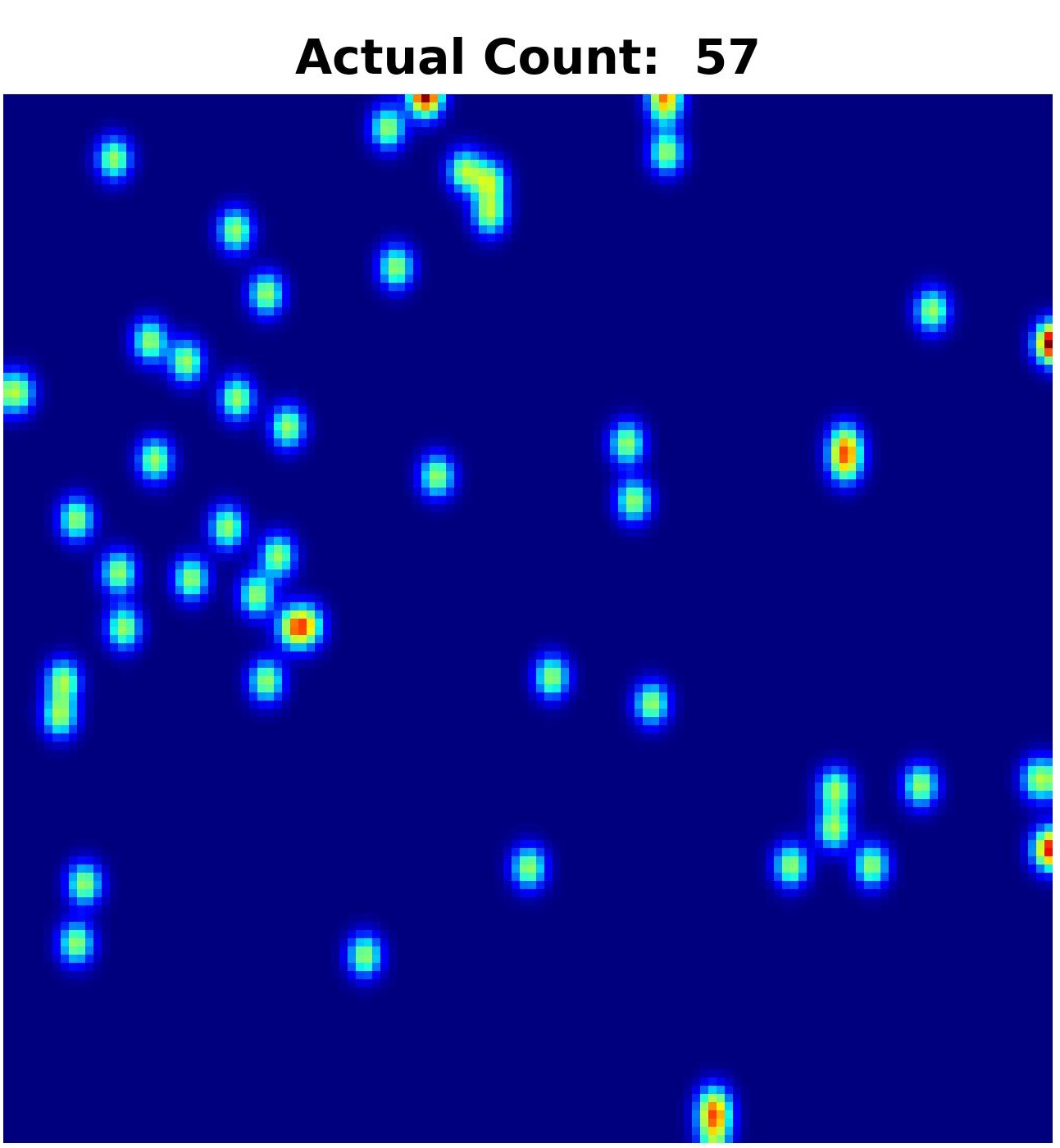} \hspace{0.1em}
    \includegraphics[width=0.4\columnwidth]{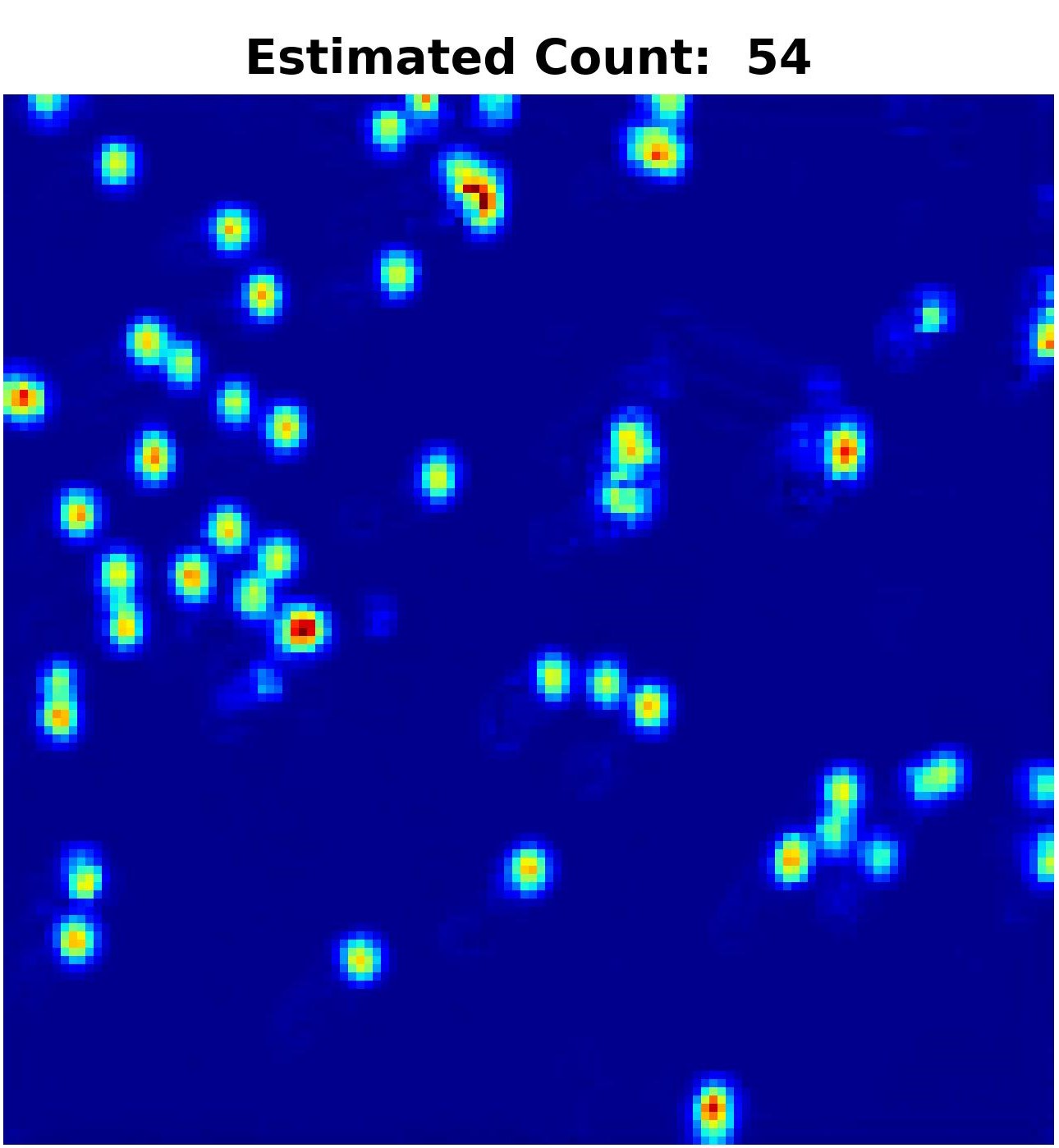}
    
    \caption{Illustration of counting prediction on a single sample: RGB image (top-left), corresponding TIR image (top-right), ground truth density map (bottom-left), and estimated density map (bottom-right).}
    \label{fig:fig1}
\end{figure}

Multimodal learning using both optical (RGB) and thermal (TIR) images for crowd counting has been proposed in recent works \cite{wu2022rgbt,liu2023rgbt,thiben2023rgbt}. The simultaneous use of both RGB and TIR images provides more information to extract enriched features from the crowd scene leading to better predictions. Nevertheless, the interest in multimodal crowd counting is increasing, there exist only a few multimodal datasets with both optical and thermal images. Thus, our aim in this paper partly is to address this challenge. We propose the use of generative adversarial networks (GANs) \cite{GANs_2014} to generate high-quality synthetic thermal images from optical images. Generative Adversarial Networks (GANs) are powerful deep learning models capable to generate realistic and high-quality data. The Pix2Pix \cite{Pix2Pix_2017, khan2023ivcnz} is a type of GAN that translate images from a given source domain to a target domain using paired training data. Pix2Pix has been successfully used in applications such as mage-to-image translation, image colorization, style transfer, etc. The generated thermal images using Pix2Pix GAN may not contain all the information that real thermal images would contain, but they certainly provide more enriched information that the model can learn. This paper thus investigates the use of Pix2Pix GANs to populate the existing monomodal RGB crowd datasets with multimodal thermal images to improve the performance of existing crowd models.

The contribution of this paper is as follows:
\begin{itemize}
\item We used a Pix2Pix GAN trained on an existing RGBT crowd dataset to generate high-quality realistic thermal images for RGB crowd images. Fig. \ref{fig:fig1} illustrates the RGB-to-TIR image translation. The efficacy of using synthetic images generated by Pix2Pix GAN in crowd counting is evaluated. For this purpose, various crowd counting models are trained on RGB, real thermal, and synthetic thermal images.
\item We design a minimal end-to-end multimodal crowd counting method (MMCount) that takes an RGB input image, generates a thermal counterpart, and trains on both RGB and thermal images to predict the crowd density map.
\item The performance of the proposed method is evaluated on three benchmark datasets using RGB-only, TIR-only, and RGB+TIR inputs to show the efficacy of the multimodal scheme.
\end{itemize}

\section{Related Work}
\subsection{Crowd Counting}
Traditional computer-vision methods were based on detecting people in crowd images using shapes, body parts, or other image information such as texture, and foreground segmentation. However, more accurate counting models today use density estimation using convolution neural networks (CNNs) and vision transformers (ViTs).
The first CNN-based crowd-counting model i.e., CrowdCNN \cite{CrowdCNN_CVPR2015} was a single-column 6-layer CNN architecture. Afterward, a large number of CNN architectures using the density estimation approach were proposed with progressive improvement in the network architecture, learning functions, and evaluation methods. MCNN \cite{MCNN_CVPR2016}, CrowdNet \cite{CrowdNet_CVPR2016}, SCNN \cite{} proposed multi-column architectures to overcome the scale variations in crowd images. MCNN uses three CNN columns with receptive fields of different sizes and combines the features learned by each column to predict the final density map.
CrowdNet \cite{CrowdNet_CVPR2016} is a 2-layer architecture containing a 5-layer deep network and a 3-layer shallow network with feature fusion for final prediction.
Scale-aware architectures replaced simple multi-column networks for learning scale variations using special multi-scale modules for scale-relevant feature extraction.
MSCNN \cite{MSCNN_ICIP2017}, a single-column network uses Inception modules \cite{Inception_CVPR2015} called multi-scale blobs (MSBs) for feature extraction in different layers.
A cascaded multi-task learning (CMTL) model \cite{CMTL_AVSS2017} is proposed to adapt to the wide variations of density levels in images. CMTL is also a two-column network. The first column is a high-level prior that classifies an input image into groups based on the total count in the image. The features learned by the high-level prior are shared with the second column that estimates the respective density map.
\par
In highly dense crowd images, simple CNN architectures such as multi-column or scale-aware models often gain limited accuracy. In highly congested scenes, deeper architectures relying on transfer learning provide better performance. Authors in \cite{CSRNet_CVPR2018} propose CSRNet \cite{CSRNet_CVPR2018} uses VGG16 \cite{VGG16_ICLR2015} as the front-end to extract features, and a CNN network with dilated convolution. Other models using transfer learning include CANNet \cite{CANNet_CVPR2019}, GSP \cite{GSP_CVPR2019}, TEDnet \cite{TEDnet_CVPR2019}, Deepcount \cite{DeepCount_ECAI2020}, SASNet \cite{SASNet_AAAI2021}, M-SFANet \cite{MSFANet-ICPR2021}, and SGANet \cite{SGANet_IEEEITS2022}.

\subsection{Generative Adversarial Networks}
The first vanilla GAN architecture was proposed in \cite{GANs_2014}. The Vanilla GAN is a two-stage architecture comprised of a generator and discriminator engaged in adversarial training. The generator creates data to fool the discriminator, while the discriminator aims to distinguish between real and generated data. During the training, the generator improves its ability to produce realistic samples. The conditional GANs (cGAN) \cite{cGAN_2014} extend the Vanilla GAN by allowing the user to generate specific types of data by providing specific conditioning information.
DCGANs \cite{DCGAN_2016} employs deep networks in both generator and discriminator to generate more realistic images.
CycleGANs \cite{CycleGAN_2017} are proposed for image-to-image translation without the need for paired training data and are used in applications such as style transfer, domain adaptation, etc.
Pix2Pix GANs \cite{Pix2Pix_2017} are based on cGANs \cite{cGAN_2014} to generate a target image conditional on a given input image.
StyleGAN \cite{StyleGAN_2021} allows a style-based generation of more realistic and diverse images.
The Pix2Pix GAN \cite{Pix2Pix_2017} provides a generic framework for image-to-image translation. Pix2Pix GANs are highly capable of translating images from one domain to another and are used in various tasks such as image colorization, style transfer, etc. Recently, Pix2Pix GANs have been used to generate synthetic near-infrared images of crops \cite{deLima2022Pix2PixNT}, lung CT scan images \cite{Toda2022LungCC}, MRA images of brain \cite{Aljohani2022}, etc. \cite{khan2023ivcnz} propose Pix2Pix GAN based image denoising architecture for fine-grained density estimation of crowd.

\section{The Proposed Method}
We propose a crowd density estimation framework MMCount. There are two essential parts; a Pix2Pix GAN, and a multimodal crowd counting network. The Pix2Pix GAN generates thermal infrared (TIR) images from optical RGB images of the crowd scene and the crowd model uses both RGB and TIR images to predict the crowd density map. The architecture of MMCount is shown in Fig. \ref{fig:prop_scheme} and is explained as follows:

\begin{figure*}
    \centering
    \includegraphics[width=0.9\textwidth]{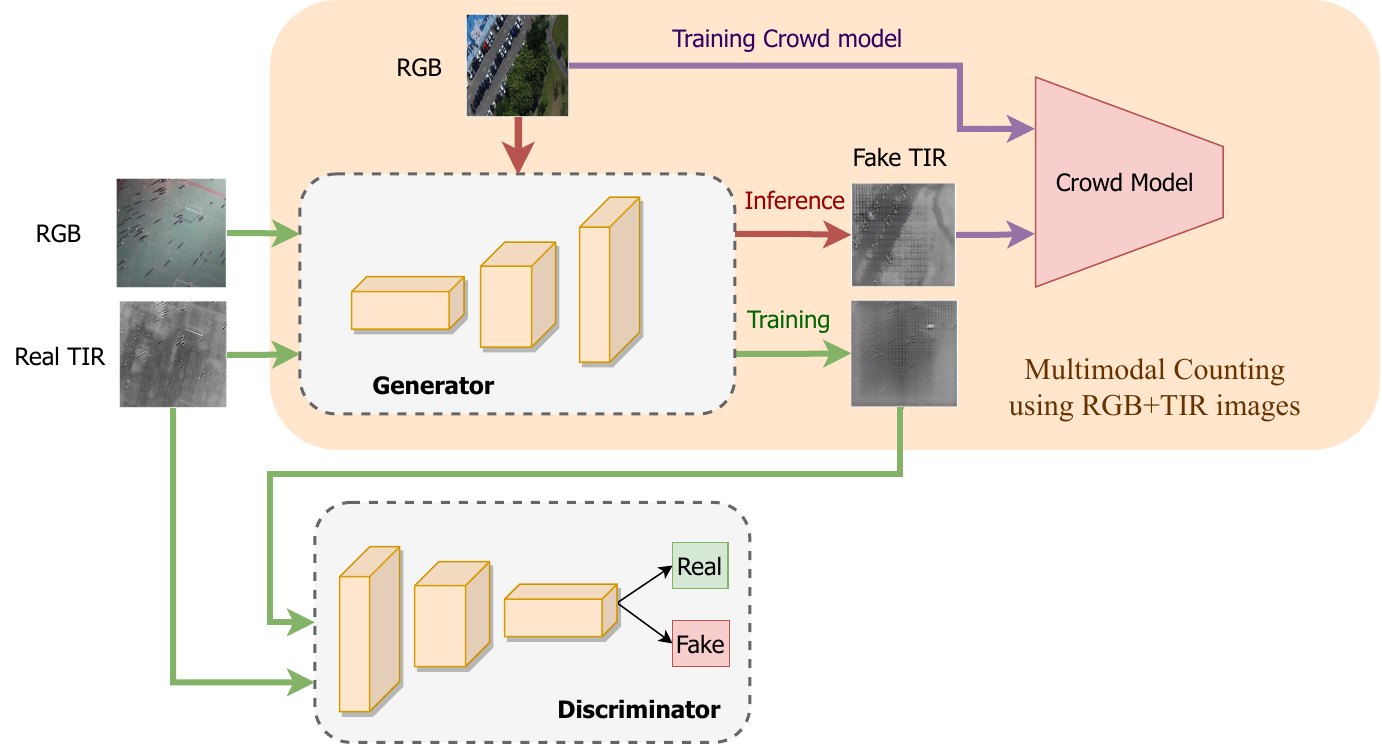}
    \caption{The proposed method for multimodal crowd counting using RGB+TIR images. The TIR images are generated by a Pix2Pix GAN trained earlier on RGB+TIR paired datasets.}
    \label{fig:prop_scheme}
\end{figure*}

\subsection{Pix2Pix GANs}
A Pix2Pix GAN consists of two main components: a generator and a discriminator.
The generator takes an input RGB image and aims to transform it into a target TIR image. The generator uses an encoder-decoder architecture. The encoder encodes the RGB image into a compact representation, and the decoder decodes this representation to generate the TIR image.

The discriminator is a CNN architecture that takes the original RGB image and the generated TIR image (by generator) and attempts to classify it as real or fake. It is trained to distinguish between the real TIR images and the TIR images generated by the generator.

During the training, the generator aims to minimize the pixel-wise $L_1$ loss to generate more realistic TIR images. The generator also aims to minimize the binary cross-entropy loss (called adversarial loss or GAN loss) to fool the discriminator. The discriminator aims to maximize the adversarial loss to correctly classify real and generated TIR images.
The pixel-wise $L_1$ loss measures the absolute pixel-wise difference between the generated output and the ground truth target and is given in Eq. \ref{eq:l1_loss}:

\begin{equation}
\mathcal{L}_{\text{L1}}(G) = \frac{1}{N} \sum_{i=1}^{N} \left| G(x_i) - y_i \right|
\label{eq:l1_loss}
\end{equation}
where $G$ is the generator network, $N$ is number of samples, $x_i$ is the input sample, and $y_i$ is the ground truth value. 

The BCE loss or the adversarial loss measures the similarity between the discriminator's predictions for the generated output and the ground truth target and is given in Eq. \ref{eq:bce_loss}:

\begin{equation}
\begin{aligned}
\mathcal{L}_{\text{BCE}}(D, G) &= - \frac{1}{N} \sum_{i=1}^{N} \bigg( y_i \log D(x_i) \\
&\quad + (1 - y_i) \log (1 - D(G(x_i))) \bigg)
\end{aligned}
\label{eq:bce_loss}
\end{equation}

where $D$ is the discriminator network, $N$ is the total number of samples, $D_{(x_i)}$ is discriminator's prediction for input $x_i$, and $D(G_{(x_i)})$ is discriminator's prediction for the generated output $G_{(x_i)}$.

\subsection{The Multimodal Counting Network}
The counting network called "MMCount" is a CNN architecture with two branches, a RGB branch, and a TIR branch. Both branches have similar structures and consist of four convolution layers (conv), each having 16, 32, 64, and 128 filters of ($3\times3$) size, respectively. Each conv layer is also followed by a Relu activation and a ($2\times2$) pooling layer with $stride=2$. The outputs of both RGB and TIR layers are concatenated and fused in the fusion layer containing 256 filters of size $3\times3$. Lastly, a $1\times1$ conv layer is used to generate the density map. The output density map generated by the MMCount is $1/8$ of the original input images (both RGB and TIR images are of the same size). The complete architecture is presented in Table \ref{tab_mmcount}

\begin{table}[!h]
    \centering
     \caption{The architecture of MMCount. The vector in Conv layers represents [input channels, output channels, kernel size, padding, stride]. Each vector in Pool layers represents [kernel size, stride].}
     \setlength{\tabcolsep}{8pt}
    \begin{tabular}{l|l|l} \toprule[0.2em]
     Type  &RGB layer      &TIR layer  \\ \midrule \midrule
     Conv  &[3,16,3,1,1]   &[1,16,3,1,1] \\
     Pool  &[2,2]          &[2,2] \\
     Conv  &[16,32,3,1,1]  &[16,32,3,1,1] \\
     Pool  &[2,2]          &[2,2] \\
     Conv  &[32,64,3,1,1]  &[32,64,3,1,1] \\
     Conv  &[64,128,3,1,1] &[64,128,3,1,1] \\ \midrule
     Conv (Fusion)  &\multicolumn{2}{c}{[256,256,3,1,1]} \\
     Conv (Regressor)  &\multicolumn{2}{c}{[256,1,1,1,1]} \\
     \bottomrule
    \end{tabular}
    \label{tab_mmcount}
\end{table}

To train the MMCount model, we use the original annotations for the RGB images which are in the form of head positions. The head positions are used to generate sparse localization maps i.e., binary matrices of pixel values (same size as the image) in which a value of 1 denotes the head position whereas a 0 represents no head. Such a dot map is used to create density maps that serve as the ground truth for the images to train the model. A density map is generated by convolving a delta function $\delta(x - x_i)$ with a Gaussian kernel $G_\sigma$, where $x_i$ are pixel values containing the head positions.

\begin{equation}
    D = \sum_{i=1}^{N}{ \delta(x-x_i) * G_\sigma}
\end{equation}
where, $N$ denotes the total number of dot points with value 1 in the dot map (i.e., total headcount in the input image). The integral of the density map $D$ is equal to the total head count in the image. Visually, this operation creates a blurring of each head annotation using the scale parameter $\sigma$. The value of $\sigma$ can be fixed \cite{SANet_ECCV2018} or adaptive \cite{CompositionLoss_2018, MSCNN_ICIP2017}.
The typical loss function used in most crowd density estimation works is the $l_2$ loss (euclidean distance) between the target and the predicted density maps and is given in Eq. \ref{eq:mse_loss}.

\begin{equation} \label{eq:mse_loss}
    L(\Theta) = \frac{1}{N} \sum_{1}^{N}{ ||D(X_i;\Theta) - D_i^{gt}||_2^2}
\end{equation}

where $N$ is the total number of samples in training data, $X_i$ is the input image, $D_i^{gt}$ is the ground truth density map, and $D(X_i;\Theta)$ is the predicted density map.

\section{Experiments and Results}

\subsection{Datasets}
\subsubsection{DroneRGBT:}
The dataset contains $3600$ pairs of RGB and thermal images. All images have a fixed resolution of $512\times640$ pixels. The images cover different scenes e.g., campus, streets, parks, parking lots, playgrounds, and plazas. The dataset is divided into a training set ($1807$ samples) and a test set ($912$ samples).

\subsubsection{ShanghaiTech Part-B:} The dataset is a large-scale crowd-counting dataset used in many studies. The dataset is split into train and test subsets consisting of 400 and 316 images, respectively. All images are of fixed size $(1024 \times 768)$.

\subsubsection{CARPK:} This dataset contains images of cars from 4 different parking lots captured using a drone (Phantom 3 Professional) at approximately 40-meter altitude. The dataset contains 1448 images split into train and test sets of sizes 989 and 459 images, respectively. The dataset contains a total of 90,000 car annotations and has been used in several object counting and object detection studies.
\par

\subsection{Evaluation metrics}
The standard and commonly used metric to evaluate the performance of crowd counting models is the mean absolute error (MAE) calculated using the following Eq. \ref{eq:mae}.

\begin{equation} \label{eq:mae}
    MAE = \frac{1}{N} \sum_{1}^{N}{\|e_n - \hat{g_n}\|}
\end{equation}


where, $N$ is the size of the dataset, $g_n$ is the target or label (actual count) and ${e_n}$ is the prediction (estimated count) in the $n^{th}$ crowd image. MAE provides per-image counting and does not take into account the incorrect density estimations in the same image. Grid Average Mean absolute Error (GAME) \cite{GAME_metric2015} can overcome these errors by computing patch-wise error i.e., an image is divided into $4^L$ non-overlapping patches and MAE is computed over each patch. GAME is thus a more robust and accurate metric for crowd-counting. It is defined in Eq. \ref{eq:game}.

\begin{equation} \label{eq:game}
    GAME = \frac{1}{N} \sum_{n=1}^{N}{ ( \sum_{l=1}^{4^L}{|e_n^l - g_n^l|)}}
\end{equation}

By setting $L=0$, GAME(0) becomes equivalent to MAE. GAME(1) means the image is divided into 4 patches and GAME(2) means 16 patches.
\subsection{Baselines}
First we evaluate the quality of generated TIR images by measuring their performance in monomodal crowd counting. For this purpose, four different models i.e., MCNN \cite{MCNN_CVPR2016}, CMTL \cite{CMTL_AVSS2017}, CSRNet \cite{CSRNet_CVPR2018}, SANet \cite{SANet_ECCV2018}, and LCDnet \cite{LCDnet_khan2023} are used.
Then, to measure the performance of MMCount in crowd density estimation using multimodal crowd counting using RGB+TIR images is evaluated. We also evaluate the counting performance of multimodal learning in MCNN and DroneNet models by feeding one CNN column with TIR images.

\subsection{Settings}
In our experiments, we used a fixed value of $\sigma$ to generate the ground truth density maps. To be more specific, a value of $7$, $10$ and $15$ are used for DroneRGBT, CARPK, and ShanghaiTech Part-B datasets, respectively.
We used the $Adam$ optimizer with a learning rate of $1 \times 10^{-3}$. We used pixel-wise $L_1$ loss function and BCE loss for training the Pix2Pix GAN and $L_2$ or MSE loss function to train all crowd counting models. The crowd counting model training terminates when the MAE error does not reduces after 10 epochs. All the models are trained on two RTX-8000 GPUs using the PyTorch framework.

\subsection{Results}
\subsection{Pix2Pix GAN Results}
\begin{figure*}
    \centering
    \begin{subfigure}[t]{0.32\textwidth}
    \includegraphics[width=0.99\textwidth]{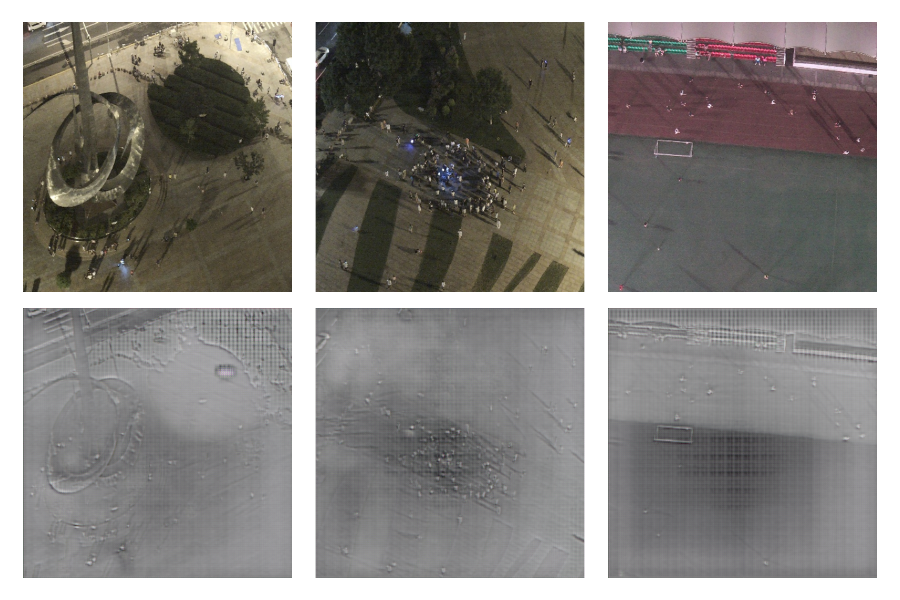}
    \caption{DroneRGBT.}
    \end{subfigure}
    \begin{subfigure}[t]{0.32\textwidth}
    \includegraphics[width=0.99\textwidth]{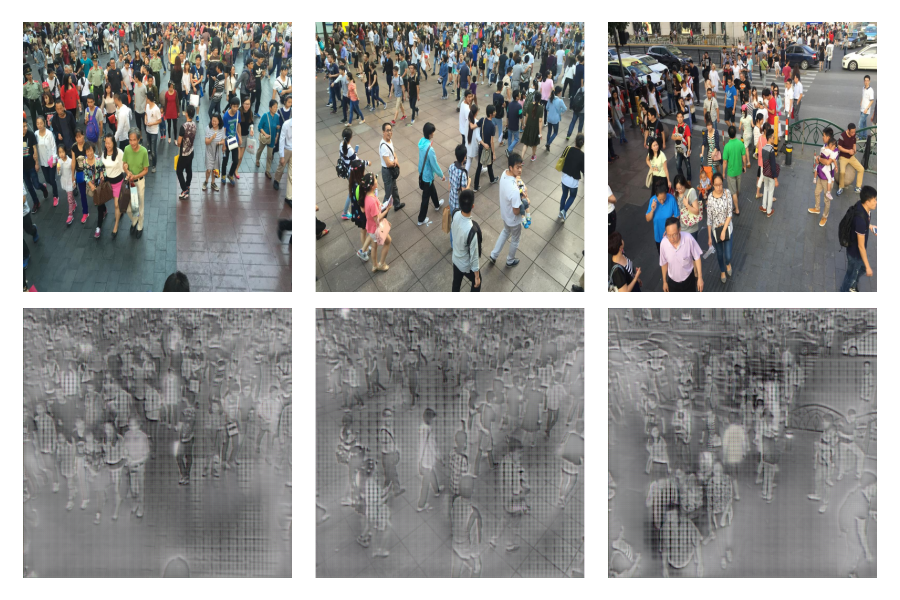}
    \caption{ShanghaiTech Part-B.}
    \end{subfigure}
    \begin{subfigure}[t]{0.32\textwidth}
    \includegraphics[width=0.99\textwidth]{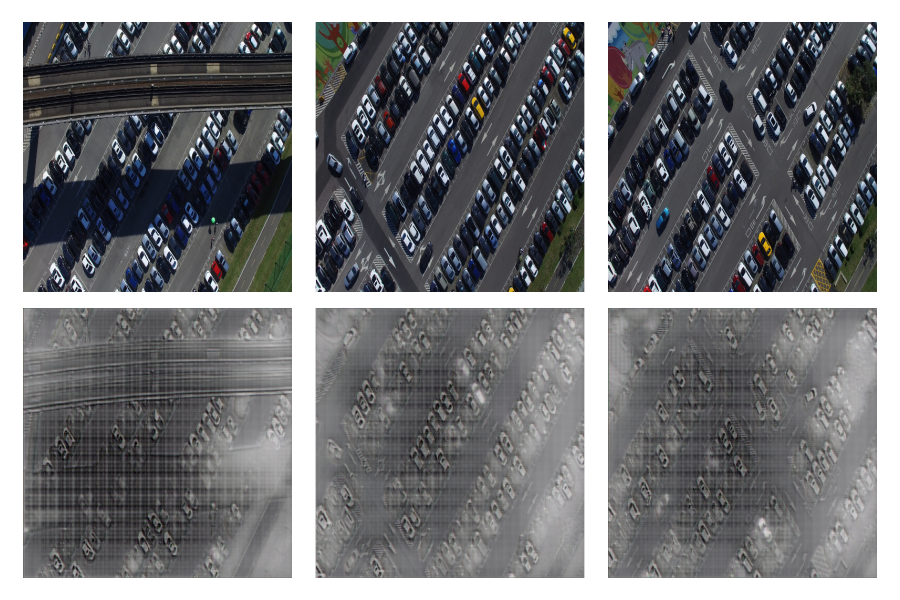}
    \caption{CARPK.}
    \end{subfigure}
    \caption{Thermal infrared (TIR) images generated using Pix2Pix GAN.}
    \label{fig:tir_pix2pix}
\end{figure*}

In the first set of experiments on Pix2Pix GANs, the synthetic TIR images are generated and stored for each dataset. The Pix2Pix model is trained on the paired RGB and TIR images provided by the DroneRGBT dataset and the trained model is then used to generate the TIR images for other datasets.
Fig. \ref{fig:tir_pix2pix} shows samples TIR images generated for the three datasets.

Interestingly, the Pix2Pix GAN architecture have been effective to generate high quality TIR images of RGB images from different datasets, despite large variations in the crowd scenes and camera settings.

\subsection{Crowd Counting Results}
To evaluate the efficacy of the generated TIR images, we trained five crowd counting models over three different inputs i.e., (i) using RGB images, (ii) using real TIR images, and (iii) using generated TIR images. The purpose is two fold: first, to understand the ability of crowd models from TIR images and second, the efficacy of generated TIR images to produce comparable results to the real TIR images. Table \ref{tab_DroneRGBT} presents the results of the first set of experiments.

\begin{table}[!h]
\centering
\caption{Accuracy comparison of crowd counting models over DroneRGBT dataset trained using (i) RGB images only versus (ii) TIR images only versus (iii) Generated TIR images only.}
\setlength{\tabcolsep}{8pt}
\renewcommand{\arraystretch}{1}

\begin{tabular}{r|ccc} \toprule[0.2em]
& \multicolumn{3}{c}{Mean Absolute Error (MAE)} \\[0.2em] \cmidrule{2-4}

\multirow{2}{*}{Model}  &RGB &TIR &Generated TIR \\[0.2em] \midrule \midrule

MCNN &17.9 &20.2 &22.5 \\[0.2em]
CMTL &18.1 &19.6 &21.4 \\[0.2em]
CSRNet &7.6 &10.8 &13.7 \\[0.2em]
SANet &16.2 &18.3 &20.8 \\[0.2em]
LCDnet &21.4 &23.2 &24.4 \\[0.2em]

\bottomrule
\end{tabular}

\label{tab_DroneRGBT}
\end{table}
It can be observed that RGB images produce higher counting accuracy (lower MAE values) due to more enriched information. The interesting information is the MAE values crowd models using generated TIR images which are only slightly lower than those of real TIR images. This strengthens our motivation to use Pix2Pix GAN to generate TIR images for a multimodal counting framework.
In the second set of experiments, we trained the proposed multimodal counting network using monomodal data (RGB-only and TIR-only images) and multimodal data (using RGB+TIR images). Furthermore, due to the multicolumn structure of MCNN and DroneNet, we extend our experiments to train these models by feeding one of three columns using the TIR images. The TIR images in both monomodal and multimodal settings in this experiment are generated by the Pix2Pix GAN. The performance is measured using the GAME metric with a value $L=0,1,2$ corresponding to the whole image, four patches per image, and 16 patches per image. The results are compared in Table \ref{tab_rgbt}.


\begin{table*}[!h]
\centering
\caption{Accuracy comparison of crowd counting models over several datasets (trained using RGB-only versus TIR-only versus RGB+TIR images). All TIR images are generated using Pix2Pix GAN. GAME(0) is equivalent to MAE.}
\setlength{\tabcolsep}{2pt}








\begin{tabular}{r|r| ccc| ccc| ccc} \toprule[0.2em]
& & \multicolumn{3}{c|}{\textbf{DroneRGBT}}  & \multicolumn{3}{c|}{\textbf{ShanghaiTech Pat-B}}   & \multicolumn{3}{c}{\textbf{CARPK}}\\ \cmidrule{3-11}

\textbf{Input} &\textbf{Model} & GAME0  &GAME1 &GAME2 & GAME0 &GAME1 &GAME2  & GAME0 &GAME1 &GAME2 \\[0.2em] \midrule \midrule

\multirow{2}{*}{MCNN}
&RGB  &17.9 &24.2 &42.0   &26.4 &34.4 &55.2    &10.1 &21.2 &43.4\\[0.2em]
&TIR  &22.5 &27.5 &51.3   &35.2 &38.2 &66.7   &16.8 &28.0 &49.0 \\[0.2em]
&RGB+TIR &\textbf{16.2} &\textbf{21.0} &\textbf{35.1}  &\textbf{23.2} &\textbf{31.5} &\textbf{48.5}  &\textbf{8.9} &\textbf{19.6} &\textbf{36.1}
 \\[0.2em] \midrule

\multirow{2}{*}{DroneNet} 
&RGB  &11.3  &22.1 &32.7   &22.4 &30.2 &41.9   &9.0 &20.5 &40.1 \\[0.2em]
&TIR  &18.6  &25.2  &40.3  &29.2 &33.4 &52.8   &15.3 &27.6 &46.2 \\[0.2em]
&RGB+TIR &\textbf{10.1} &\textbf{18.8} &\textbf{28.4}   &\textbf{20.0} &\textbf{29.7} &\textbf{38.3}   &\textbf{8.1} &\textbf{18.5} &\textbf{26.8} \\[0.2em] \midrule

\multirow{2}{*}{MMCount} 
&RGB     &10.8  &21.1  &32.4   &21.6 &28.2 &40.1   &8.2 &18.3 &37.5 \\[0.2em]
&TIR     &16.0  &23.3  &40.6   &27.7 &33.6 &49.9   &15.0 &25.6 &42.8 \\[0.2em]
&RGB+TIR &\textbf{9.2}   &\textbf{18.0}  &\textbf{26.0}   &\textbf{18.2} &\textbf{28.0} &\textbf{36.4}   &\textbf{7.8} &\textbf{15.2} &\textbf{25.0} \\[0.2em] \bottomrule
\end{tabular}

\label{tab_rgbt}
\end{table*}

\section{Conclusions}
This paper addresses the challenge of the limited availability of training data in crowd counting scenarios and proposes the use of generative adversarial networks to instantly generate scene-specific multimodal data. A Pix2Pix GAN is used to generate thermal infrared images for the available color RGB images. The multimodal data (RGB+TIR) is then used to train or fine-tune a crowd counting model. Experiments are conducted using three publicly available datasets. The results show improvements in model performance when trained using actual RGB and synthetic TIR images. The TIR images are generated using a Pix2Pix GAN trained on cross-scene drone images and applied to new and unseen RGB images. We believe the results can be further improved by training the GAN model on images of more correlated scenarios. As a future work, we plan to develop novel lightweight GAN architectures for real-time performance.


\section*{Acknowledgement}
This publication was made possible by the PDRA award PDRA7-0606-21012 from the Qatar National Research Fund (a member of The Qatar Foundation). The statements made herein are solely the responsibility of the authors.

\bibliographystyle{apalike}
{\small
\bibliography{biblio}}  

\end{document}